\documentclass{article}

\PassOptionsToPackage{numbers, compress}{natbib}

\newif\ifpreprint
\preprinttrue   
\ifpreprint
  \usepackage[preprint]{neurips_2026}
\else
  \usepackage{neurips_2026}
\fi

\usepackage[utf8]{inputenc} 
\usepackage[T1]{fontenc}    
\usepackage{hyperref}       
\usepackage{url}            
\usepackage{booktabs}       
\usepackage{amsfonts}       
\usepackage{nicefrac}       
\usepackage{microtype}      
\usepackage{xcolor}         
\usepackage{comment}
\usepackage{graphicx}
\usepackage{adjustbox}
\usepackage{amsmath}
\usepackage{wrapfig}
\usepackage[table]{xcolor}

\definecolor{bestcolor}{RGB}{255, 200, 200}      
\definecolor{secondcolor}{RGB}{255, 230, 180}    

\title{WildSplatter: Feed-forward 3D Gaussian Splatting with Appearance Control from Unconstrained Images}

\author{%
  Yuki Fujimura$^1$, Takahiro Kushida$^2$, Kazuya Kitano$^1$, Takuya Funatomi$^3$, Yasuhiro Mukaigawa$^1$\\
  $^1$NAIST, $^2$Ritsumeikan University, $^3$Kyoto University\\
  \texttt{\{fujimura.yuki,kitano.kazuya,mukaigawa\}@is.naist.jp} \\
  \texttt{tkushida@fc.ritsumei.ac.jp} \quad
  \texttt{funatomi.takuya.2c@kyoto-u.ac.jp}
}

\begin{document}

\maketitle

\begin{figure}[h]
\centering
\includegraphics[width=1.\textwidth]{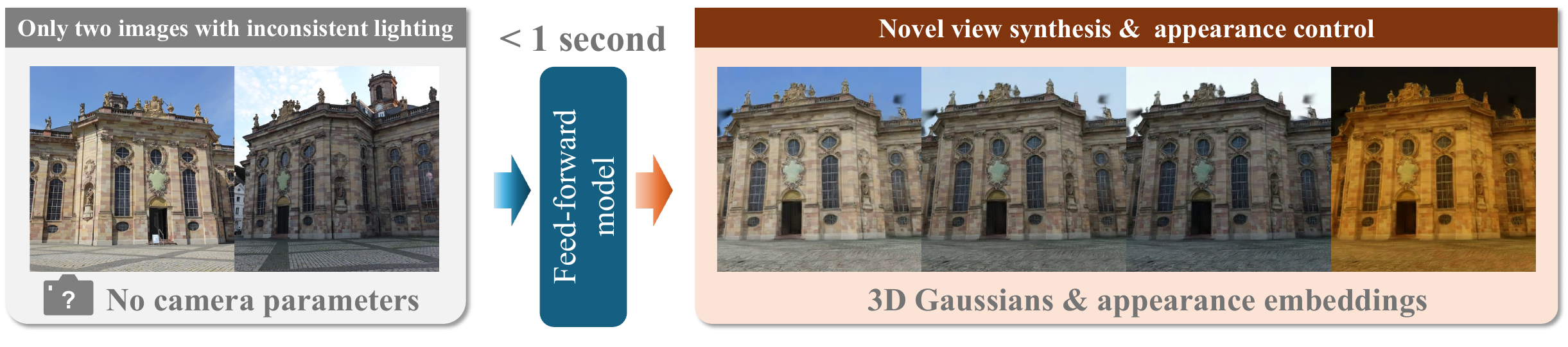}
\caption{We propose WildSplatter, a feed-forward 3DGS model for unconstrained images with unknown camera parameters and varying lighting conditions. Our method reconstructs 3D Gaussians and appearance embeddings from sparse input views in under one second, while also enabling appearance control under diverse lighting conditions.}
\label{fig:teaser}
\end{figure}

\begin{abstract}
  We propose WildSplatter, a feed-forward 3D Gaussian Splatting (3DGS) model for unconstrained images with unknown camera parameters and varying lighting conditions. 3DGS is an effective scene representation that enables high-quality, real-time rendering; however, it typically requires iterative optimization and multi-view images captured under consistent lighting with known camera parameters. WildSplatter is trained on unconstrained photo collections and jointly learns 3D Gaussians and appearance embeddings conditioned on input images. This design enables flexible modulation of Gaussian colors to represent significant variations in lighting and appearance. Our method reconstructs 3D Gaussians from sparse input views in under one second, while also enabling appearance control under diverse lighting conditions. Experimental results demonstrate that our approach outperforms existing pose-free 3DGS methods on challenging real-world datasets with varying illumination.
  \ifpreprint
  Project page: \url{https://github.com/yfujimura/WildSplatter}
  \fi
\end{abstract}

\section{Introduction}

3D Gaussian Splatting (3DGS)~\cite{Kerbl2023} has recently emerged as an effective method for representing complex 3D scenes using collections of volumetric Gaussian primitives.
In this framework, Gaussian parameters are optimized to reproduce a set of multi-view images given known camera poses.
Once trained, the representation enables photorealistic novel view synthesis via efficient differentiable rasterization.
Owing to its high rendering quality and efficiency, 3DGS has strong potential across various domains, including virtual and augmented reality, robotics, and 3D content creation, where realistic and efficient scene representations are essential.

However, the original 3DGS framework requires iterative optimization for each scene, which is computationally expensive and time-consuming.
To address this limitation, feed-forward 3DGS models~\cite{Charatan2024,Chen2024,Chen2024mvsplat360} have been proposed.
These models take sparse input views and predict pixel-aligned 3D Gaussians, enabling fast inference without iterative optimization.
More recently, pose-free feed-forward 3DGS models~\cite{Jiang2025,Smart2024,Ye2025,Zhang2025} have further relaxed the requirement for known camera parameters by eliminating the need for extrinsic, or even intrinsic, parameters, thereby improving applicability in real-world scenarios.

Another limitation of the original 3DGS framework is its assumption that input images are captured under consistent lighting conditions.
Although 3DGS models scene appearance using spherical harmonics to account for view-dependent effects, it struggles to represent large appearance variations caused by changes in lighting across images.
A typical example is Internet photo collections in outdoor scenes, which are captured with different cameras and under diverse conditions (e.g., time of day or weather), leading to significant appearance variation.
Extending 3DGS to such unconstrained image collections is therefore an important and challenging problem.
Prior work has attempted to address this issue by introducing appearance embeddings optimized per view and per Gaussian~\cite{Dahmani2024,Kulhanek2024}, or by injecting features from reference images~\cite{Jiacong2024,Zhang2024_GS-W}.
However, these approaches still rely on iterative optimization, limiting their efficiency.

In this paper, we propose a feed-forward 3DGS model for unconstrained images with unknown camera parameters and varying lighting conditions, termed {\bf WildSplatter} (Fig.~\ref{fig:teaser}).
The proposed model is trained on in-the-wild photo collections~\cite{Tung2024} to estimate 3D Gaussians in a feed-forward manner.
To account for appearance variations across input images, our approach jointly learns 3D Gaussians and appearance embeddings conditioned on the input images.
The learned appearance embeddings enable flexible modulation of Gaussian colors, allowing the model to better handle variations in lighting and appearance.
Furthermore, we demonstrate that the learned embeddings generalize across scenes, enabling controllable appearance under diverse lighting conditions.
Experimental results show that our method outperforms existing pose-free 3DGS approaches on challenging real-world datasets with varying illumination.

Our contributions are summarized as follows:
\begin{itemize}
\item We propose WildSplatter, a pose-free feed-forward 3DGS model that handles unconstrained images with varying lighting conditions.
\item We introduce a joint learning framework of 3D Gaussians and appearance embeddings for flexible appearance modeling.
\item We demonstrate that the learned embeddings generalize across scenes, enabling appearance interpolation and cross-dataset appearance control.
\item We show that our method outperforms existing pose-free 3DGS methods on challenging outdoor scene datasets.
\end{itemize}
\section{Related work}

\subsection{Feed-forward 3D Gaussian Splatting}

Feed-forward 3DGS models take input images and directly estimate 3D Gaussians, enabling fast inference without time-consuming iterative optimization.
These models typically predict pixel-aligned 3D Gaussians.
pixelSplat~\cite{Charatan2024} proposes probabilistic, differentiable Gaussian generation along camera rays, while MVSplat~\cite{Chen2024} employs plane-sweep cost volumes computed from known camera parameters to estimate Gaussian centers.

Recently, 3D vision foundation models such as DUSt3R~\cite{Wang2024dust3r} and MASt3R~\cite{Leroy2024} have been proposed, enabling 3D geometry estimation from only two input images without requiring camera parameters.
Building on these approaches, Splatt3R~\cite{Smart2024} and NoPoSplat~\cite{Ye2025} incorporate 3D Gaussian prediction heads, enabling pose-free feed-forward 3D Gaussian estimation.
Such pose-free models further enhance applicability in real-world scenarios.
FreeSplatter~\cite{Xu2024} can handle an arbitrary number of input views without requiring camera parameters.
In these approaches, pixel-aligned Gaussians also enable camera pose recovery via PnP solvers~\cite{Hartley2003}.

FLARE~\cite{Zhang2025} adopts a cascade design in which camera poses are first estimated and subsequently refined, followed by 3D Gaussian estimation.
PF3Splat~\cite{Hong2025} and SPFSplat~\cite{Huang2025} similarly estimate camera poses beforehand and use them for Gaussian prediction, removing the need for ground-truth camera poses during training.
AnySplat~\cite{Jiang2025} proposes a differentiable voxelization module to reduce pixel-aligned Gaussians, improving scalability with respect to the number of input images.
Depth Anything 3~\cite{depthanything3} is a foundation model that estimates depth maps from an arbitrary number of input images and can also be adapted for 3D Gaussian estimation.

\subsection{Novel-view synthesis from unconstrained images}

Novel view synthesis from unconstrained outdoor images with varying lighting conditions is a challenging task, as it requires modeling significant appearance variations across input images.
NeRF-W~\cite{Martin2021} is the first to address this problem within the neural radiance fields (NeRF) framework~\cite{Mildenhall2020}.
While the original NeRF represents scene radiance as a multi-layer perceptron (MLP) conditioned on position and viewing direction, NeRF-W introduces learnable per-image appearance embeddings to model appearance variations.
Ha-NeRF~\cite{Chen2022} predicts appearance embeddings using a convolutional neural network instead of optimizing them directly, enabling cross-dataset generalization.
CR-NeRF~\cite{Yang2023_CR-NeRF} leverages local grid features for improved appearance estimation.
NeRF-OSR~\cite{Rudnev2022} enables explicit relighting by incorporating spherical harmonics-based illumination conditioned on surface normals, viewing directions, and shadows.

3DGS~\cite{Kerbl2023} has also been extended to handle unconstrained images, enabling real-time and high-quality rendering compared to NeRF-based methods.
SWAG~\cite{Dahmani2024} introduces learnable per-image appearance embeddings that are fed into an MLP together with Gaussian attributes to determine final colors.
WildGaussians~\cite{Kulhanek2024} estimates affine transformation parameters using learnable embeddings defined per view and per Gaussian.
Wild-GS~\cite{Jiacong2024} queries triplane features constructed from reference images to extract local appearance information.
GS-W~\cite{Zhang2024_GS-W} learns adaptive feature sampling from reference images for improved detail reconstruction.
In contrast to these methods with implicit appearance modeling, LumiGauss~\cite{Kaleta2025} extends 2D Gaussian Splatting~\cite{Huang2024} to inverse rendering with spherical harmonics-based shadow modeling.
However, all of these approaches assume known camera parameters and rely on iterative optimization.

WildCAT3D~\cite{Alper2025} is also related to our work, as it is a generative framework trained on unconstrained image collections.
However, our method differs from WildCAT3D in several key aspects:
(1) WildCAT3D primarily focuses on generating multi-view images from a single input image;
(2) it does not enforce geometric consistency through explicit 3D representations such as NeRF or 3D Gaussians; and
(3) its reliance on diffusion-based sampling limits real-time performance.
In contrast, our method estimates 3D Gaussians from sparse views in a feed-forward manner and enables real-time novel view synthesis.
\section{Method}

\subsection{Overview}

\begin{figure}[t]
\centering
\includegraphics[width=1.0\textwidth]{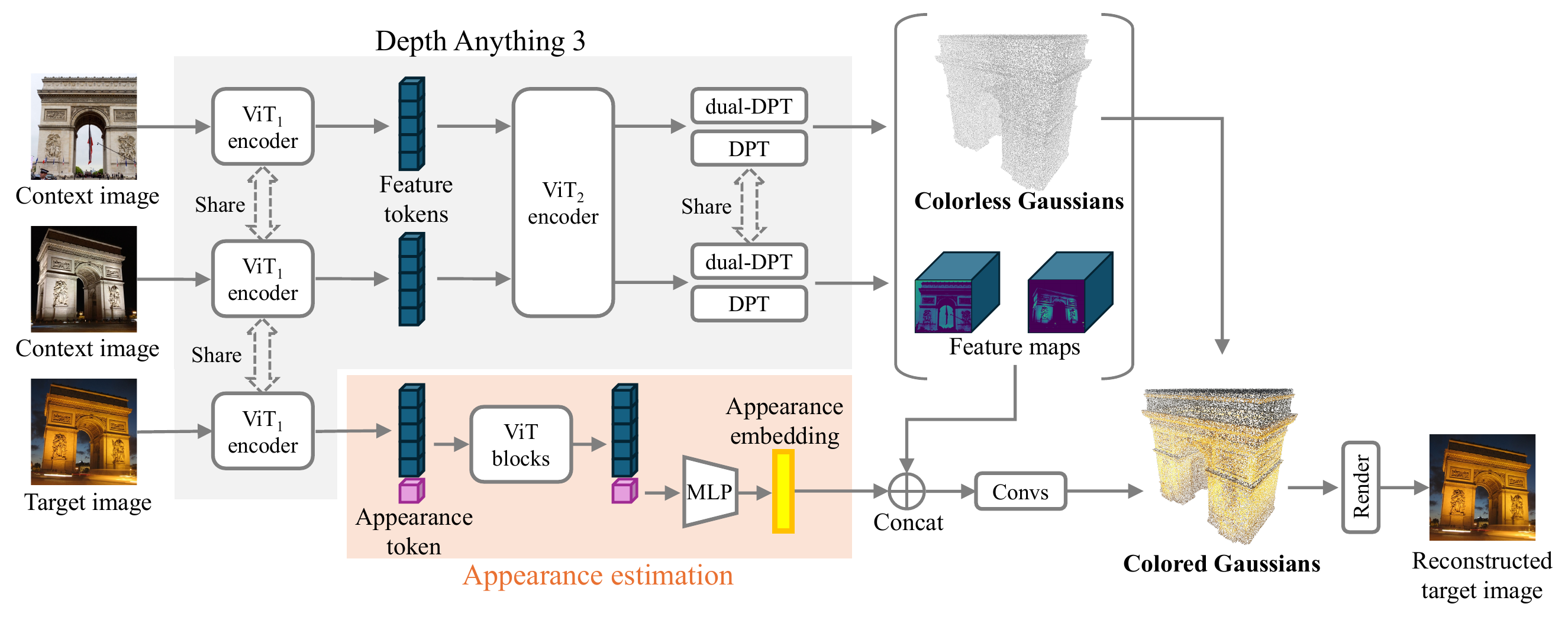}
\caption{
Overview of the proposed method.
Given sparse-view context images captured under inconsistent lighting, the model first estimates colorless 3D Gaussian geometry and local scene features.
During training, target images are additionally used to extract global appearance embeddings, which are combined with local features to predict Gaussian colors.
This design enables disentangled modeling of geometry and appearance for robust rendering under varying lighting conditions.
}
\label{fig:overview}
\end{figure}

We propose a pose-free, feed-forward 3DGS model trained on unconstrained image collections.
Figure~\ref{fig:overview} provides an overview of the proposed model.
The model is trained on in-the-wild photo collections~\cite{Tung2024} to estimate 3D Gaussians in a feed-forward manner.

To address large appearance variations across input images, our model explicitly disentangles geometry and appearance.
We first estimate the geometry of 3D Gaussians without colors, and then predict their appearance, based on the assumption that geometry remains consistent under varying lighting conditions.
Specifically, the model takes sparse-view context images captured under inconsistent lighting conditions as input.
These images are used to estimate colorless 3D Gaussian geometry and local scene features.
During training, target images are additionally fed into the model to estimate global appearance embeddings.
These embeddings are combined with local scene features to predict the colors of 3D Gaussians.
The colored 3D Gaussians are then used to render the target images, from which the training loss is computed.
This design is motivated by the observation that scene geometry remains largely invariant under lighting changes, while appearance variations can be effectively captured in a low-dimensional latent space (see the supplementary material for empirical validation).

\subsection{Architecture}

\paragraph{Vision Transformer backbone}
The architecture of our model is based on Depth Anything 3~\cite{depthanything3}, which consists of a Vision Transformer (ViT) encoder based on DINOv2~\cite{oquab2023} and DPT heads~\cite{Ranftl2021} for estimating Gaussian primitives.
The ViT encoder is divided into two parts, $\text{ViT}_1$ and $\text{ViT}_2$.
The former includes local (intra-frame) attention to extract frame-level features, while the latter incorporates global (inter-frame) attention to aggregate features across frames.

Let the input context images be $\{\mathbf{I}^c_i\}_{i=1}^{N_c}$ and the target images be $\{\mathbf{I}^t_j\}_{j=1}^{N_t}$, where $N_c$ and $N_t$ denote the numbers of context and target images, respectively.
Let the image resolution be $H\times W$, i.e., $\mathbf{I}^c_i, \mathbf{I}^t_j \in \mathbb{R}^{H\times W \times 3}$.
All images are first fed into the Transformer backbone as
\begin{equation}
\mathbf{T}^c_i = \text{ViT}_1(\mathbf{I}^c_i), \quad
\mathbf{T}^t_j = \text{ViT}_1(\mathbf{I}^t_j),
\end{equation}
where each image is processed independently by frame-level self-attention blocks, producing context and target feature tokens, respectively.
The context feature tokens are then fed into subsequent Transformer blocks as
\begin{equation}
\{\mathcal{T}_i\}_{i=1}^{N_c} = \text{ViT}_2(\{\mathbf{T}^c_i\}_{i=1}^{N_c}),
\end{equation}
where local and global attention are alternated, following the mechanism used in VGGT~\cite{Wang2025}.
Each $\mathcal{T}_i$ denotes a set of feature tokens obtained from different attention blocks.

\paragraph{Geometry estimation of 3D Gaussians}
We first estimate the geometry of 3D Gaussians (centers, opacities, rotations, and scales) from context images, based on the assumption that geometry remains invariant under lighting changes.

The feature tokens from the context images are fed into a dual-DPT head~\cite{depthanything3} to estimate depth maps $\{\mathbf{D}_i\}_{i=1}^{N_c}$ and ray maps $\{[\mathbf{o}_i,\mathbf{d}_i]\}_{i=1}^{N_c}$:
\begin{equation}
\mathbf{D}_i, [\mathbf{o}_i,\mathbf{d}_i] = \text{dual-DPT}(\mathcal{T}_i),
\end{equation}
where $\mathbf{D}_i \in \mathbb{R}^{H \times W \times 1}$ and $[\mathbf{o}_i,\mathbf{d}_i] \in \mathbb{R}^{H \times W \times 6}$ consists of ray origins $\mathbf{o}_i \in \mathbb{R}^{H \times W \times 3}$ and directions $\mathbf{d}_i \in \mathbb{R}^{H \times W \times 3}$.
In addition, Gaussian parameters are estimated via a DPT head as
\begin{equation}
\{\boldsymbol{\alpha}_i, \mathbf{r}_i, \mathbf{s}_i, \Delta\mathbf{D}_i, \mathbf{f}_i\} = \text{DPT}(\mathcal{T}_i),
\end{equation}
where $\boldsymbol{\alpha}_i \in \mathbb{R}^{H \times W \times 1}$, $\mathbf{r}_i \in \mathbb{R}^{H \times W \times 4}$, and $\mathbf{s}_i \in \mathbb{R}^{H \times W \times 3}$ denote opacity, rotation, and scale, respectively.
$\Delta\mathbf{D}_i$ represents depth offsets for better alignment with object surfaces.
The Gaussian centers are computed as
$\boldsymbol{\mu}_i = \mathbf{o}_i + (\mathbf{D}_i + \Delta \mathbf{D}_i)\mathbf{d}_i$.

We also extract feature maps $\mathbf{f}_i \in \mathbb{R}^{H\times W \times d_l}$, where $d_l$ denotes the channel dimension, before the final convolutional layer. These feature maps encode local scene information for appearance estimation.

\paragraph{Appearance estimation of 3D Gaussians}
Although the original 3DGS framework models view-dependent color using spherical harmonics, it struggles to represent large appearance variations across unconstrained images with varying lighting conditions.
To address this limitation, we estimate a global appearance embedding for each target image to modulate Gaussian colors during rendering.

As shown in Fig.~\ref{fig:overview}, a learnable appearance token is concatenated with the feature tokens of each target image $\mathbf{T}_j^t$ and processed by shallow Transformer blocks.
The resulting token is passed through a shallow MLP to produce an appearance embedding $\mathbf{e}_j \in \mathbb{R}^{d_g}$, where $d_g$ denotes the embedding dimension, representing the global appearance of $\mathbf{I}_j$.

To render $\mathbf{I}_j$, we condition Gaussian colors on $\mathbf{e}_j$.
The embedding is spatially broadcast as $\hat{\mathbf{e}}_j \in \mathbb{R}^{H\times W\times d_g}$ and concatenated with local feature maps $\{\mathbf{f}_i\}_{i=1}^{N_c}$.
These are fed into convolutional layers to estimate Gaussian colors:
\begin{equation}
\mathbf{c}_i^j = \text{Convs}(\mathbf{f}_i \oplus \hat{\mathbf{e}}_j),
\end{equation}
where $\oplus$ denotes channel-wise concatenation and $\text{Convs}$ consists of two convolutional layers.
$\mathbf{c}_i^j \in \mathbb{R}^{H \times W \times k}$ denotes spherical harmonics coefficients.
Finally, we obtain 3D Gaussians for each target image as
\begin{equation}
\{\mathbf{G}_i^j\}_{i=1}^{N_c} = \{ \boldsymbol{\mu}_i, \boldsymbol{\alpha}_i, \mathbf{r}_i, \mathbf{s}_i, \mathbf{c}_i^j \}_{i=1}^{N_c}.
\end{equation}
For all target images, we generate $\{\{\mathbf{G}_i^j\}_{i=1}^{N_c}\}_{j=1}^{N_t}$ and render each target image independently.

\begin{figure}[t]
\centering
\includegraphics[width=1.0\textwidth]{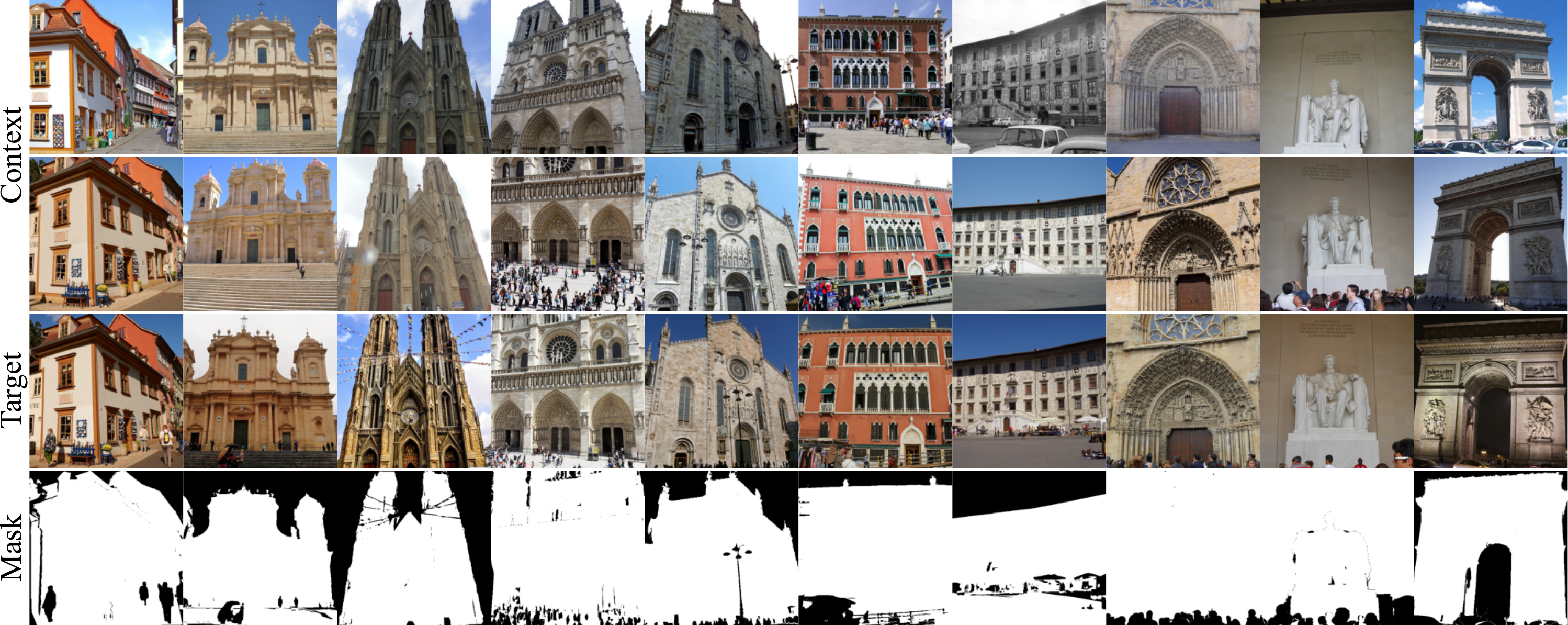}
\caption{
Examples of training samples.
The first two rows show context images captured under different lighting conditions, while the third row shows the corresponding target images.
The last row shows visibility masks computed by warping depth maps from the target views to the context views.
}
\label{fig:dataset}
\end{figure}

\subsection{Training}

\paragraph{Training dataset}\label{sec:training_dataset}
Our model is trained on a collection of unconstrained images.
Specifically, we use the MegaScenes dataset~\cite{Tung2024}, which consists of Internet photos of famous landmarks worldwide.
Although each scene contains multi-view images with camera parameters estimated by COLMAP~\cite{Schonberger2016}, the images are unstructured, requiring the construction of view sets with sufficient overlap for training.

Following prior work~\cite{Alper2025, Tung2024}, we first estimate monocular depth maps for all images using Depth Anything 3.
These depth maps are aligned with COLMAP sparse depth by estimating scale and shift via RANSAC.
By warping depth maps across views, we compute overlap regions and select image pairs with sufficient overlap as training samples.
More details are provided in the supplementary material.

Since 3DGS cannot render regions that are not visible from the context views, we enforce visibility constraints during training.
Moreover, in-the-wild images in MegaScenes often contain transient objects (e.g., pedestrians), which violate multi-view consistency.
To address these issues, we generate visibility masks based on depth consistency.
Let $\widetilde{\mathbf{D}}_j^t$ denote the scale-and-shift aligned depth map of the $j$-th target view, and $\{ \widetilde{\mathbf{D}}_i^c \}_{i=1}^{N_c}$ denote the aligned depth maps of the context views.
We warp the depth map of the target view to the context views and compute the visibility mask for the $j$-th target view as
\begin{equation}\label{eq:mask}
\mathbf{M}_j(p) = 
\begin{cases}
1 & \text{if } \exists i \in \{1,\dots,N_c\} \text{ s.t. }
| \log \widetilde{\mathbf{D}}_i^c(\pi_{j \to i}(p)) - \log \widetilde{\mathbf{D}}_{j \to i}^t(p) | < \delta, \\
0 & \text{otherwise},
\end{cases}
\end{equation}
where $\pi_{j \to i}(p)$ denotes the projection of pixel $p$ in the $j$-th target view onto the $i$-th context view, and $\widetilde{\mathbf{D}}_{j \to i}^t(p)$ denotes the depth value $\widetilde{\mathbf{D}}_j^t(p)$ after transforming the corresponding 3D point into the camera coordinate system of the $i$-th context view. Here, $\delta$ is a threshold.

Figure~\ref{fig:dataset} shows examples of training samples consisting of two context images and one target image with visibility masks.
The visibility masks are used in the training loss to suppress the effects of unobserved regions and transient objects.
Note that the masks are applied only to the target views.
We empirically observe that the model learns to suppress transient objects in the context views by reducing their opacities during training.
More details are provided in the supplementary material.

\paragraph{Training loss}
During training, we render the target images and compute a loss.
However, the dataset and the model outputs have inconsistent scales, requiring alignment of the estimated 3D Gaussians to the dataset scale.
We estimate scale and translation between point clouds derived from $\{\mathbf{D}_i\}_{i=1}^{N_c}$ and $\{ \widetilde{\mathbf{D}}_i^c\}_{i=1}^{N_c}$ during training.
To account for ambiguity and noise in monocular depth estimation, we employ a weighted least-squares formulation, where weights are computed based on multi-view consistency of $\{ \widetilde{\mathbf{D}}_i^c\}_{i=1}^{N_c}$.
Details are provided in the supplementary material.

The training loss consists of pixel-wise mean squared error (MSE) and perceptual loss based on LPIPS~\cite{Zhang2018}:
\begin{equation}
\mathcal{L} = \sum_{j=1}^{N_t} 
\text{MSE}(\mathbf{M}^s_j \odot \mathbf{I}^t_j, \mathbf{M}^s_j \odot \hat{\mathbf{I}}^t_j) 
+ \lambda \, \text{LPIPS}(\mathbf{M}^s_j \odot \mathbf{I}^t_j, \mathbf{M}^s_j \odot \hat{\mathbf{I}}^t_j),
\end{equation}
where $\hat{\mathbf{I}}_j^t$ is the rendered target image, and $\odot$ denotes element-wise multiplication.
$\mathbf{M}^s_j$ is a visibility mask extended with sky regions to explicitly model sky appearance.
The sky regions are obtained from sky probability maps predicted by Depth Anything 3.
$\lambda$ controls the weight of the LPIPS loss.

\begin{table}
\caption{Quantitative comparison for novel-view synthesis on the NeRF-OSR dataset~\cite{Rudnev2022}.
We report PSNR ($\uparrow$) and LPIPS ($\downarrow$).
The best results are highlighted in red, and the second-best in orange.}
\label{tab:quantitative_comparison}
\centering
\begin{adjustbox}{width=\textwidth}
\begin{tabular}{lcc|cc|cc|cc|cc|cc}
\toprule
Scene & \multicolumn{2}{c}{{\it europa}} & \multicolumn{2}{c}{{\it lk2}} & \multicolumn{2}{c}{{\it lwp}} & \multicolumn{2}{c}{{\it schloss}} & \multicolumn{2}{c}{{\it st}} & \multicolumn{2}{c}{{\it stjohann}} \\
\bottomrule
\multicolumn{13}{c}{2 views} \\
& PSNR & LPIPS & PSNR & LPIPS & PSNR & LPIPS & PSNR & LPIPS & PSNR & LPIPS & PSNR & LPIPS  \\
\midrule
WildGaussians & \cellcolor{secondcolor}15.00 & \cellcolor{secondcolor}0.454 & 14.19 & 0.434 & \cellcolor{secondcolor}13.14 & \cellcolor{secondcolor}0.519 & \cellcolor{bestcolor}17.51 & \cellcolor{secondcolor}0.385 & 11.02 & \cellcolor{secondcolor}0.509 & 12.18 & \cellcolor{secondcolor}0.396 \\
SPFSplat & 10.98 & 0.574 & 12.59 & 0.573 & 10.17 & 0.615 & 13.05 & 0.464 & 12.40 & 0.529 & 10.32 & 0.524 \\
AnySplat & 13.52 & 0.461 & 12.40 & 0.525 & 10.75 & 0.592 & 13.39 & 0.485 & 10.93 & 0.534 & 11.38 & 0.535 \\
Depth Anything 3 & 14.31 & 0.461 & \cellcolor{secondcolor}14.54 & \cellcolor{secondcolor}0.428 & 12.29 & 0.600 & 16.03 & 0.463 & \cellcolor{secondcolor}13.99 & 0.547 & \cellcolor{secondcolor}12.19 & 0.476 \\
\textbf{WildSplatter} & \cellcolor{bestcolor}16.00 & \cellcolor{bestcolor}0.421 & \cellcolor{bestcolor}16.37 & \cellcolor{bestcolor}0.375 & \cellcolor{bestcolor}13.81 & \cellcolor{bestcolor}0.455 & \cellcolor{secondcolor}17.41 & \cellcolor{bestcolor}0.334 & \cellcolor{bestcolor}14.59 & \cellcolor{bestcolor}0.437 & \cellcolor{bestcolor}13.66 & \cellcolor{bestcolor}0.384 \\
\bottomrule
\multicolumn{13}{c}{3 views} \\
& PSNR & LPIPS & PSNR & LPIPS & PSNR & LPIPS & PSNR & LPIPS & PSNR & LPIPS & PSNR & LPIPS  \\
\midrule
WildGaussians & \cellcolor{secondcolor}13.15 & 0.505 & 14.73 & 0.448 & 11.50 & 0.637 & \cellcolor{secondcolor}15.37 & 0.473 & 13.08 & \cellcolor{bestcolor}0.392 & 13.27 & \cellcolor{secondcolor}0.386 \\
SPFSplat & 11.33 & 0.615 & 9.89 & 0.681 & 10.71 & 0.669 & 11.29 & 0.557 & 13.31 & 0.509 & 11.88 & 0.602 \\
AnySplat & 11.98 & 0.517 & 13.40 & 0.503 & 10.00 & 0.642 & 12.33 & 0.507 & 12.18 & 0.485 & 12.18 & 0.523 \\
Depth Anything 3 & 13.11 & \cellcolor{secondcolor}0.480 & \cellcolor{secondcolor}15.56 & \cellcolor{secondcolor}0.431 & \cellcolor{secondcolor}12.58 & \cellcolor{secondcolor}0.587 & 14.90 & \cellcolor{secondcolor}0.432 & \cellcolor{secondcolor}14.97 & 0.411 & \cellcolor{secondcolor}13.32 & 0.462 \\
\textbf{WildSplatter} & \cellcolor{bestcolor}15.87 & \cellcolor{bestcolor}0.421 & \cellcolor{bestcolor}16.92 & \cellcolor{bestcolor}0.404 & \cellcolor{bestcolor}13.01 & \cellcolor{bestcolor}0.527 & \cellcolor{bestcolor}17.55 & \cellcolor{bestcolor}0.342 & \cellcolor{bestcolor}15.05 & \cellcolor{secondcolor}0.393 & \cellcolor{bestcolor}16.20 & \cellcolor{bestcolor}0.373 \\
\bottomrule
\multicolumn{13}{c}{4 views} \\
& PSNR & LPIPS & PSNR & LPIPS & PSNR & LPIPS & PSNR & LPIPS & PSNR & LPIPS & PSNR & LPIPS  \\
\midrule
WildGaussians & \cellcolor{secondcolor}14.23 & 0.489 & 15.05 & 0.399 & 12.30 & 0.573 & 14.43 & 0.461 & 12.77 & \cellcolor{bestcolor}0.399 & \cellcolor{secondcolor}15.00 & \cellcolor{bestcolor}0.347 \\
SPFSplat & 11.92 & 0.602 & 13.83 & 0.600 & 7.19 & 0.728 & 12.61 & 0.526 & 13.18 & 0.536 & 11.15 & 0.667 \\
AnySplat & 11.84 & 0.537 & 13.35 & 0.520 & 10.40 & 0.598 & 12.39 & 0.513 & 12.37 & 0.495 & 13.04 & 0.501 \\
Depth Anything 3 & 13.24 & \cellcolor{secondcolor}0.468 & \cellcolor{secondcolor}16.33 & \cellcolor{secondcolor}0.394 & \cellcolor{secondcolor}12.89 & \cellcolor{secondcolor}0.528 & \cellcolor{secondcolor}15.40 & \cellcolor{secondcolor}0.413 & \cellcolor{bestcolor}15.14 & 0.455 & 14.72 & 0.423 \\
\textbf{WildSplatter} & \cellcolor{bestcolor}15.99 & \cellcolor{bestcolor}0.421 & \cellcolor{bestcolor}17.72 & \cellcolor{bestcolor}0.369 & \cellcolor{bestcolor}13.95 & \cellcolor{bestcolor}0.482 & \cellcolor{bestcolor}17.20 & \cellcolor{bestcolor}0.350 & \cellcolor{secondcolor}14.83 & \cellcolor{secondcolor}0.419 & \cellcolor{bestcolor}16.91 & \cellcolor{secondcolor}0.363 \\
\bottomrule
\end{tabular}
\end{adjustbox}
\end{table}

\begin{figure}[t]
\centering
\includegraphics[width=0.85\textwidth]{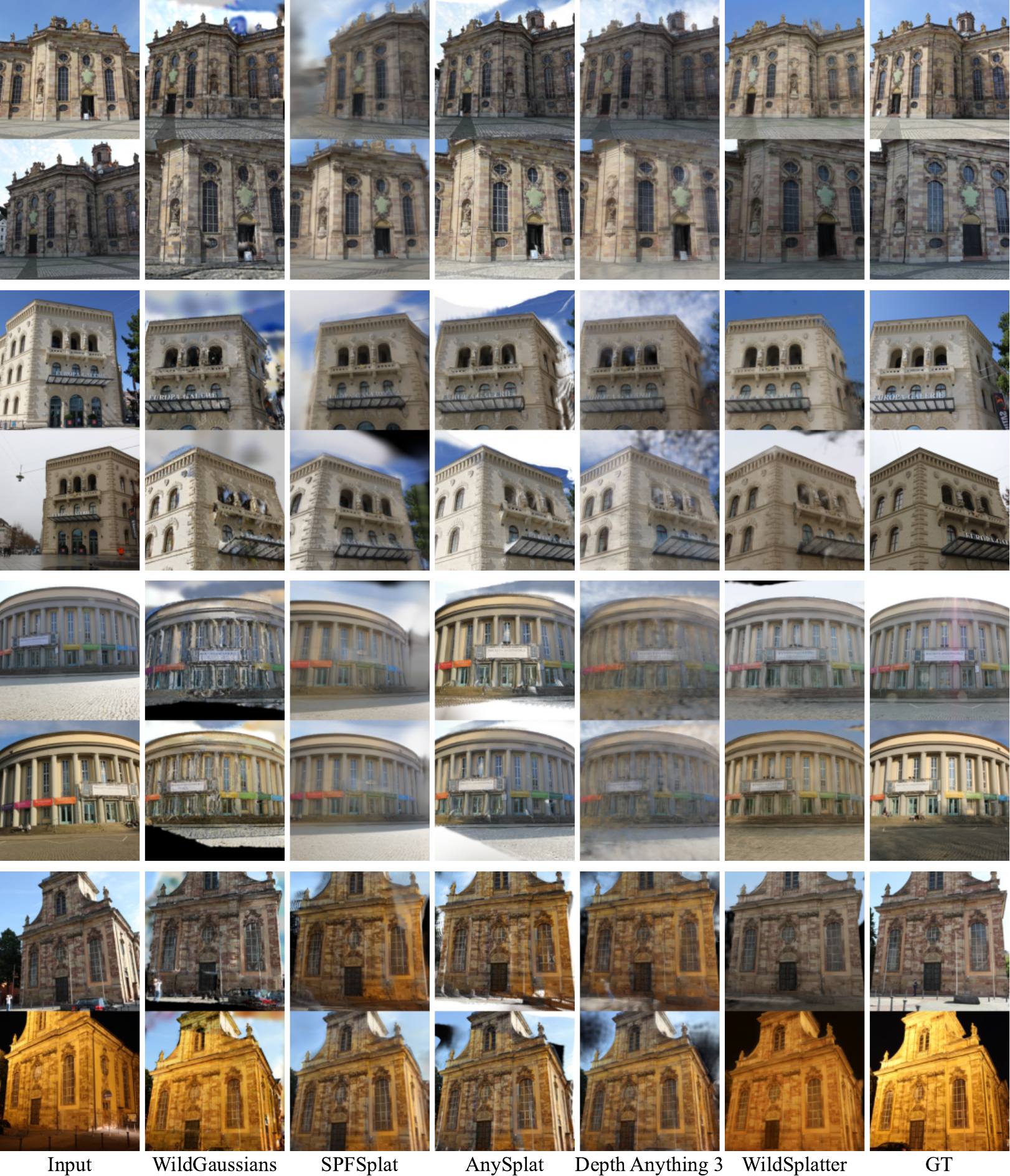}
\caption{
Qualitative comparison for novel-view synthesis on the NeRF-OSR dataset~\cite{Rudnev2022}.
}
\label{fig:qualitative_comparison}
\end{figure}

\section{Experiments}

\subsection{Experimental setup}

\paragraph{Evaluation dataset}

We evaluate our method on the NeRF-OSR dataset~\cite{Rudnev2022}.
Each scene consists of outdoor image sequences captured on different dates under varying lighting conditions.
For context views, we select images captured on different dates.
For novel-view synthesis evaluation, we select four target images captured on the same date as each context view, ensuring consistent lighting conditions.

\paragraph{Implementation details}

Our model is initialized with a pretrained Depth Anything 3~\cite{depthanything3}.
To leverage its strong geometric priors, we freeze the backbone ViT and the dual-DPT head used for depth and ray estimation.
We train the DPT head for Gaussian parameter prediction and the appearance estimation module, which includes ViT blocks, an MLP, and output convolutional layers.
The resolution of input images is $504 \times 504$.
Additional implementation details, including training hyperparameters, are provided in the supplementary material.

\subsection{Experimental results}

\paragraph{Baseline methods}

We primarily compare our WildSplatter with state-of-the-art pose-free feed-forward 3DGS models, including SPFSplat~\cite{Huang2025}, AnySplat~\cite{Jiang2025}, and Depth Anything 3~\cite{depthanything3}.
For SPFSplat and AnySplat, we use and render images with the resolution of $256 \times 256$ and $448 \times 448$, respectively, to match their training settings.
These methods model Gaussian colors using standard spherical harmonics; thus, we evaluate our appearance modeling under unconstrained imaging conditions.


We additionally compare our method with WildGaussians~\cite{Kulhanek2024}, an optimization-based 3DGS approach with appearance modeling for unconstrained images.
Such optimization-based methods generally struggle with sparse inputs.
For a fair comparison, we initialize the 3D Gaussians using depth maps predicted by Depth Anything 3.
Noisy depth values are filtered using predicted confidence maps.

For WildGaussians and our WildSplatter, we use appearance embeddings for the target images estimated from the context images on the same dates.

\begin{figure}[t]
\centering
\includegraphics[width=0.9\textwidth]{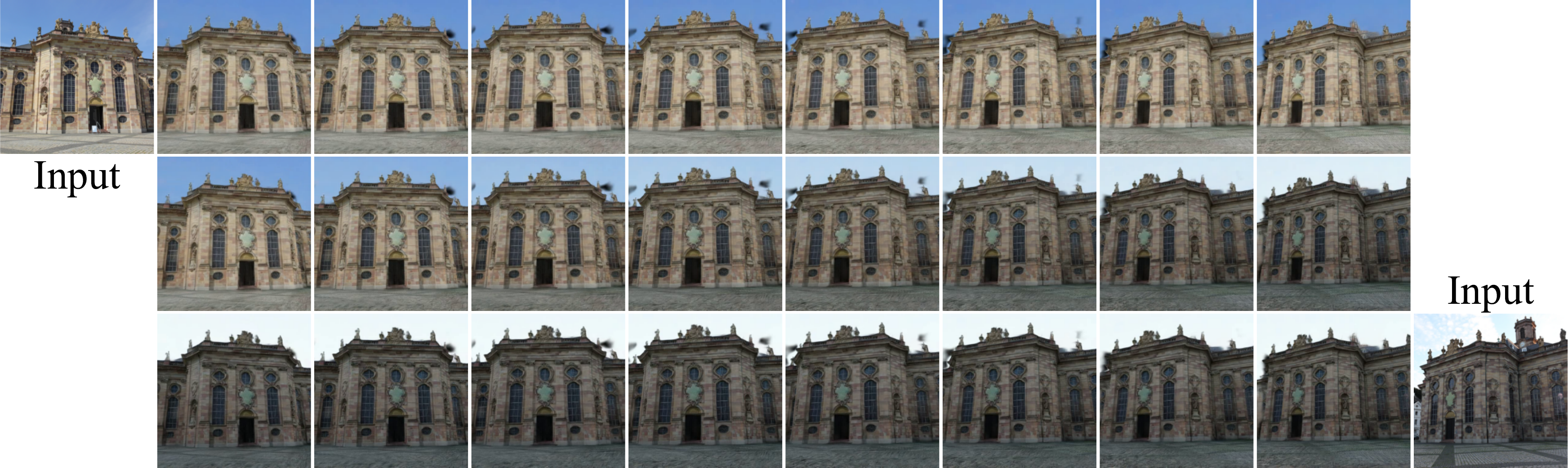}
\caption{Appearance interpolation. The first and third rows show rendered images from novel views with fixed appearance embeddings from the two input views, respectively. The second row shows rendered images with linearly interpolated appearance embeddings.}
\label{fig:app_interp}
\end{figure}

\paragraph{Results}

Table~\ref{tab:quantitative_comparison} presents quantitative results for novel-view synthesis on the NeRF-OSR dataset.
We evaluate performance using two to four context images per scene.
We report PSNR and LPIPS as evaluation metrics.

As shown in the table, our method outperforms competing methods across most metrics and scenes under different numbers of input context views.
Notably, increasing the number of input images does not always improve performance for all methods.
This is because the target images differ depending on the number of context views in our evaluation protocol (four target images per context image), and the context images are selected from different dates with varying lighting conditions, making the task more challenging.

Figure~\ref{fig:qualitative_comparison} presents qualitative comparison with the input of two context views.
For the compared feed-forward 3DGS without explicit appearance modeling, the rendered novel-views tend to contain mixed appearance.
On the other hand, our method correctly model the appearance variation under the unconstrained images.

\paragraph{Limitations}
Despite the effectiveness of our approach, our method models appearance using a single global embedding, which can lead to slight color drift and limited ability to represent complex lighting effects such as shadows, as shown in Fig.~\ref{fig:qualitative_comparison}.
Addressing these limitations is an important direction for future work, for example by incorporating more expressive appearance representations or explicit inverse rendering models.

\subsection{Discussion}

\paragraph{Appearance interpolation}
Our learned appearance embeddings enable flexible appearance control of 3D Gaussians.
Figure~\ref{fig:app_interp} presents examples of simultaneous novel-view synthesis and appearance interpolation.
Given two context views, we estimate the corresponding appearance embeddings.
The first and third rows show view interpolation under each appearance embedding, while the second row shows joint view and appearance interpolation obtained by linearly interpolating the embeddings.
These results demonstrate the smoothness of the learned embedding space and its effectiveness for appearance interpolation.

\begin{figure}[t]
\centering
\includegraphics[width=0.8\textwidth]{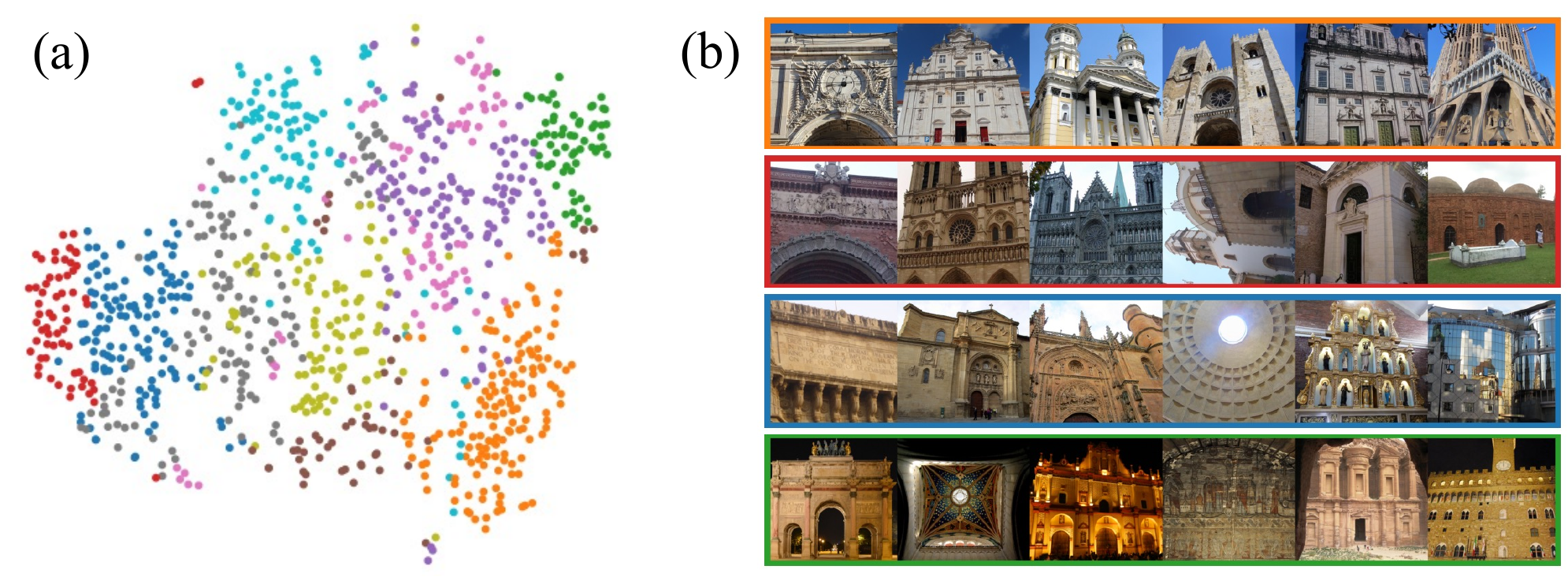}
\caption{
(a) t-SNE visualization~\cite{Vandermaaten2008} of learned appearance embeddings, where each point corresponds to an image and colors indicate K-means clusters.
(b) Representative images for each cluster, showing that images within the same cluster share similar appearance characteristics.
}
\label{fig:app_embed_vis}
\end{figure}

\paragraph{Visualization of learned appearance embeddings}
To further analyze the learned embedding space, Fig.~\ref{fig:app_embed_vis} shows a t-SNE visualization~\cite{Vandermaaten2008} of appearance embeddings for a subset of the training images
, showing that visually similar images are grouped together in the embedding space.

\begin{figure}[t]
\centering
\includegraphics[width=0.8\textwidth]{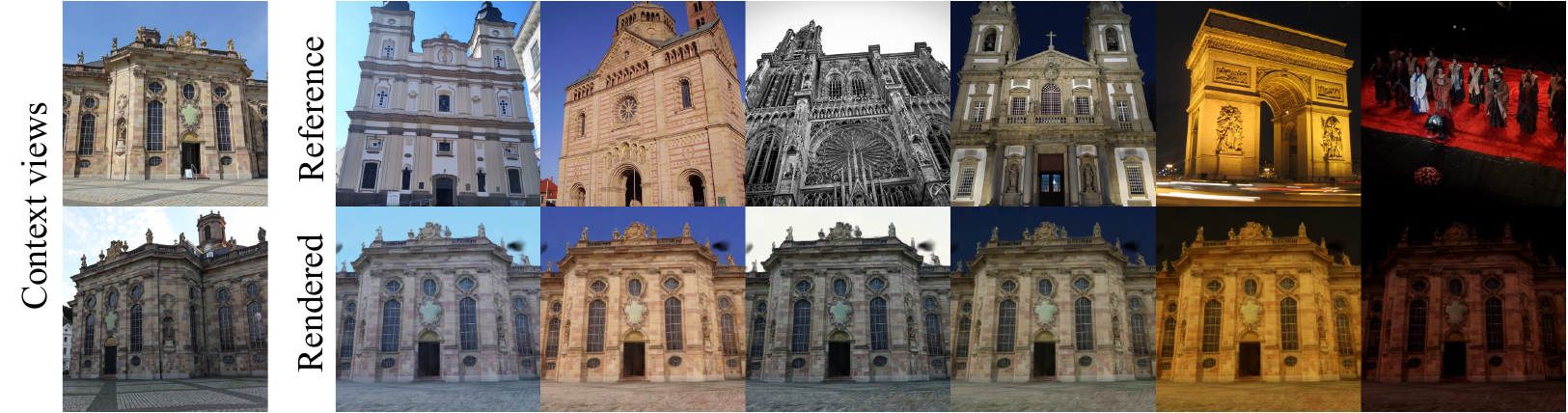}
\caption{
Cross-dataset appearance control.
The geometry of 3D Gaussians estimated from context views is kept fixed, while their colors are modulated using appearance embeddings extracted from reference images of different scenes.
}
\label{fig:app_ctrl}
\end{figure}

\paragraph{Cross-dataset appearance control}
The results in Fig.~\ref{fig:app_embed_vis} suggest that appearance embeddings can generalize across scenes.
Figure~\ref{fig:app_ctrl} demonstrates cross-scene appearance control, where the colors of 3D Gaussians estimated from context views are modulated using appearance embeddings extracted from reference images of different scenes.

\paragraph{Runtime}
\begin{wraptable}{r}{0.45\linewidth}
\vspace{-20pt}
\centering
\caption{Runtime comparison}
\label{tab:runtime_comparison}
\begin{tabular}{lc}
\toprule
Method & Runtime \\
\midrule
WildGaussians~\cite{Kulhanek2024} & 1.5 min \\
Depth Anything 3~\cite{depthanything3} & 0.368 s \\
WildSplatter & 0.375 s \\
\bottomrule
\end{tabular}
\vspace{-10pt}
\end{wraptable}

Table~\ref{tab:runtime_comparison} reports runtime comparisons with two input views.
All runtimes are measured on a single NVIDIA RTX 6000 Ada GPU.
Compared to the optimization-based method~\cite{Kulhanek2024}, our method significantly reduces computation time.
Furthermore, our runtime is comparable to that of Depth Anything 3~\cite{depthanything3}, indicating that the additional appearance estimation module does not introduce a significant computational overhead.

\section{Conclusion}

In this paper, we proposed WildSplatter, a feed-forward 3DGS model for unconstrained images with unknown camera parameters and varying lighting conditions, enabling fast inference from sparse input views.
Our model jointly learns 3D Gaussians and appearance embeddings, and we demonstrated that the learned embeddings effectively modulate Gaussian colors, enabling appearance interpolation and cross-dataset appearance control.
These results highlight the effectiveness of our approach in handling both geometric reconstruction and appearance variation in real-world scenarios.

\ifpreprint
\begin{ack}
This work was supported by JSPS KAKENHI Grant Number 26K02931.
\end{ack}
\fi

\bibliographystyle{plain}
\bibliography{main}

@String(CVPR= {IEEE Conf. Comput. Vis. Pattern Recog.})

@String(ICCV= {Int. Conf. Comput. Vis.})

@String(ECCV= {Eur. Conf. Comput. Vis.})

@String(NIPS= {Adv. Neural Inform. Process. Syst.})

@String(TOG= {ACM Trans. Graph.})

@String(ICLR = {Int. Conf. Learn. Represent.})

@String(CVPR  = {CVPR})

@String(ICCV  = {ICCV})

@String(ECCV  = {ECCV})

@String(NIPS  = {NeurIPS})

@String(TOG   = {ACM TOG})

@String(ICLR  = {ICLR})

@String(WACV = {WACV})

@String(ICML = {ICML})

@String(SIGGRAPH = {ACM SIGGRAPH})

@InProceedings{Schonberger2016,
author = {Schonberger, Johannes L. and Frahm, Jan-Michael},
title = {Structure-From-Motion Revisited},
booktitle = CVPR,
year = {2016}
}

@InProceedings{Zhang2018,
author = {Zhang, Richard and Isola, Phillip and Efros, Alexei A. and Shechtman, Eli and Wang, Oliver},
title = {The Unreasonable Effectiveness of Deep Features as a Perceptual Metric},
booktitle = CVPR,
year = {2018}
}

@InProceedings{Ranftl2021,
    author    = {Ranftl, Ren\'e and Bochkovskiy, Alexey and Koltun, Vladlen},
    title     = {Vision Transformers for Dense Prediction},
    booktitle = ICCV,
    year      = {2021},
}

@inproceedings{Mildenhall2020,
 author = {Ben Mildenhall and Pratul P. Srinivasan and Matthew Tancik and Jonathan T. Barron and Ravi Ramamoorthi and Ren Ng},
 title = {NeRF: Representing Scenes as Neural Radiance Fields for View Synthesis},
 booktitle = ECCV,
 year = 2020
}

@article{Kerbl2023,
author = {Kerbl, Bernhard and Kopanas, Georgios and Leimkuehler, Thomas and Drettakis, George},
title = {3D Gaussian Splatting for Real-Time Radiance Field Rendering},
year = {2023},
volume = {42},
number = {4},
journal = TOG,
articleno = {139},
numpages = {14},
}

@InProceedings{Charatan2024,
    author    = {Charatan, David and Li, Sizhe Lester and Tagliasacchi, Andrea and Sitzmann, Vincent},
    title     = {pixelSplat: 3D Gaussian Splats from Image Pairs for Scalable Generalizable 3D Reconstruction},
    booktitle = CVPR,
    year      = {2024},
    pages     = {19457-19467}
}

@InProceedings{Chen2024,
author="Chen, Yuedong
and Xu, Haofei
and Zheng, Chuanxia
and Zhuang, Bohan
and Pollefeys, Marc
and Geiger, Andreas
and Cham, Tat-Jen
and Cai, Jianfei",
title="MVSplat: Efficient 3D Gaussian Splatting from Sparse Multi-view Images",
booktitle=ECCV,
year=2024,
pages="370--386",
}

@InProceedings{Chen2024mvsplat360,
    title     = {MVSplat360: Feed-Forward 360 Scene Synthesis from Sparse Views},
    author    = {Chen, Yuedong and Zheng, Chuanxia and Xu, Haofei and Zhuang, Bohan and Vedaldi, Andrea and Cham, Tat-Jen and Cai, Jianfei},
    booktitle = NIPS,
    year      = {2024},
}

@inproceedings{Xu2024,
  title={FreeSplatter: Pose-free Gaussian Splatting for Sparse-view 3D Reconstruction},
  author={Jiale Xu and Shenghua Gao and Ying Shan},
  booktitle = ICCV,
  pages     = {25442-25452},
  year={2025}
}

@InProceedings{Wang2024dust3r,
    author    = {Wang, Shuzhe and Leroy, Vincent and Cabon, Yohann and Chidlovskii, Boris and Revaud, Jerome},
    title     = {DUSt3R: Geometric 3D Vision Made Easy},
    booktitle = CVPR,
    year      = {2024},
    pages     = {20697-20709}
}

@InProceedings{Leroy2024,
author    = {Vincent Leroy, Yohann Cabon, J\'{e}r\^{o}me Revaud},
title ={Grounding Image Matching in 3D with MASt3R},
booktitle = {arXiv preprint arXiv:2406.09756},
year={2024}
}

@InProceedings{Smart2024,
author    = {Brandon Smart and Chuanxia Zheng and Iro Laina and Victor Adrian Prisacariu
},
title ={Splatt3R: Zero-shot Gaussian Splatting from Uncalibrated Image Pairs},
booktitle = {arXiv preprint arXiv:2408.13912},
year={2024}
}

@inproceedings{Ye2025,
  title={No Pose, No Problem: Surprisingly Simple 3D Gaussian Splats from Sparse Unposed Images},
  author={Botao Ye and Sifei Liu and Haofei Xu and Xueting Li and Marc Pollefeys and Ming-Hsuan Yang and Songyou Peng},
  booktitle = ICLR,
  year={2025}
}

@inproceedings{Loshchilov2019,
  title={Decoupled Weight Decay Regularization},
  author={Loshchilov, Ilya and Hutter, Frank},
  booktitle=ICLR,
  year={2019},
}

@book{Hartley2003,
author = {Hartley, Richard and Zisserman, Andrew},
title = {Multiple View Geometry in Computer Vision},
year = {2003},
publisher = {Cambridge University Press},
}

@inproceedings{Huang2024,
author = {Huang, Binbin and Yu, Zehao and Chen, Anpei and Geiger, Andreas and Gao, Shenghua},
title = {2D Gaussian Splatting for Geometrically Accurate Radiance Fields},
year = {2024},
booktitle = SIGGRAPH
}

@InProceedings{Zhang2025,
    author    = {Zhang, Shangzhan and Wang, Jianyuan and Xu, Yinghao and Xue, Nan and Rupprecht, Christian and Zhou, Xiaowei and Shen, Yujun and Wetzstein, Gordon},
    title     = {FLARE: Feed-forward Geometry, Appearance and Camera Estimation from Uncalibrated Sparse Views},
    booktitle = CVPR,
    month     = {June},
    year      = {2025},
    pages     = {21936-21947}
}

@InProceedings{Hong2025,
      title   = {PF3plat: Pose-Free Feed-Forward 3D Gaussian Splatting},
      author  = {Sunghwan Hong and Jaewoo Jung and Heeseong Shin and Jisang Han and Jiaolong Yang and Chong Luo and Seungryong Kim},
      booktitle = ICML,
      year    = {2025}
    }

@InProceedings{Martin2021,
    author    = {Martin-Brualla, Ricardo and Radwan, Noha and Sajjadi, Mehdi S. M. and Barron, Jonathan T. and Dosovitskiy, Alexey and Duckworth, Daniel},
    title     = {NeRF in the Wild: Neural Radiance Fields for Unconstrained Photo Collections},
    booktitle = CVPR,
    year      = {2021},
    pages     = {7210-7219}
}

@inproceedings{Kulhanek2024,
  title={WildGaussians: 3D Gaussian Splatting in the Wild},
  author={Kulhanek, Jonas and Peng, Songyou and Kukelova, Zuzana and Pollefeys, Marc and Sattler, Torsten},
  booktitle=NIPS,
  year={2024}
}

@InProceedings{depthanything3,
  title={Depth Anything 3: Recovering the visual space from any views},
  author={Haotong Lin and Sili Chen and Jun Hao Liew and Donny Y. Chen and Zhenyu Li and Guang Shi and Jiashi Feng and Bingyi Kang},
  booktitle = ICLR,
  year={2026}
}

@inproceedings{Tung2024,
      title={MegaScenes: Scene-Level View Synthesis at Scale}, 
      author={Tung, Joseph and Chou, Gene and Cai, Ruojin and Yang, Guandao and Zhang, Kai and Wetzstein, Gordon and Hariharan, Bharath and Snavely, Noah},
      booktitle=ECCV,
      year={2024}
    }

@InProceedings{Rudnev2022,
      title={NeRF for Outdoor Scene Relighting},
      author={Viktor Rudnev and Mohamed Elgharib and William Smith and Lingjie Liu and Vladislav Golyanik and Christian Theobalt},
      booktitle=ECCV,
      year={2022}
}

@inproceedings{Zhang2024_GS-W,
  title={Gaussian in the wild: 3d gaussian splatting for unconstrained image collections},
  author={Zhang, Dongbin and Wang, Chuming and Wang, Weitao and Li, Peihao and Qin, Minghan and Wang, Haoqian},
  booktitle=ECCV,
  pages={341--359},
  year={2024},
}

@InProceedings{Chen2022,
    author    = {Chen, Xingyu and Zhang, Qi and Li, Xiaoyu and Chen, Yue and Feng, Ying and Wang, Xuan and Wang, Jue},
    title     = {Hallucinated Neural Radiance Fields in the Wild},
    booktitle = CVPR,
    year      = {2022},
    pages     = {12943-12952}
}

@InProceedings{Yang2023_CR-NeRF,
    author    = {Yang, Yifan and Zhang, Shuhai and Huang, Zixiong and Zhang, Yubing and Tan, Mingkui},
    title     = {Cross-Ray Neural Radiance Fields for Novel-View Synthesis from Unconstrained Image Collections},
    booktitle = ICCV,
    year      = {2023},
    pages     = {15901-15911}
}

@InProceedings{Kaleta2025,
    author    = {Kaleta, Joanna and Kania, Kacper and Trzcinski, Tomasz and Kowalski, Marek},
    title     = {LumiGauss: Relightable Gaussian Splatting in the Wild},
    booktitle = WACV,
    year      = {2025},
    pages     = {1-10}
}

@InProceedings{Alper2025,
  title={WildCAT3D: Appearance-Aware Multi-View Diffusion in the Wild},
  author={Alper, Morris and Novotny, David and Kokkinos, Filippos and Averbuch-Elor, Hadar and Monnier, Tom},
  booktitle=NIPS,
  year={2025}
}

@article{Jiang2025,
  title={Anysplat: Feed-forward 3d gaussian splatting from unconstrained views},
  author={Jiang, Lihan and Mao, Yucheng and Xu, Linning and Lu, Tao and Ren, Kerui and Jin, Yichen and Xu, Xudong and Yu, Mulin and Pang, Jiangmiao and Zhao, Feng and others},
  journal=TOG,
  volume={44},
  number={6},
  pages={1--16},
  year={2025}
}

@InProceedings{Huang2025,
    author    = {Huang, Ranran and Mikolajczyk, Krystian},
    title     = {No Pose at All: Self-Supervised Pose-Free 3D Gaussian Splatting from Sparse Views},
    booktitle = ICCV,
    year      = {2025},
    pages     = {27947-27957}
}

@article{oquab2023,
  title   = {DINOv2: Learning Robust Visual Features without Supervision},
  author  = {Oquab, Maxime and Darcet, Timoth{\'e}e and Moutakanni, Th{\'e}o and Vo, Huy and Szafraniec, Marc and Khalidov, Vasil and Fernandez, Pierre and Haziza, Daniel and Massa, Francisco and El-Nouby, Alaaeldin and Assran, Mahmoud and Ballas, Nicolas and Galuba, Wojciech and Howes, Russell and Huang, Po-Yao and Li, Shang-Wen and Misra, Ishan and Rabbat, Michael and Sharma, Vasu and Synnaeve, Gabriel and Xu, Hu and Jegou, Herv{\'e} and Mairal, Julien and Labatut, Patrick and Joulin, Armand and Bojanowski, Piotr},
  journal = {arXiv preprint arXiv:2304.07193},
  year    = {2023}
}

@InProceedings{Wang2025,
    author    = {Wang, Jianyuan and Chen, Minghao and Karaev, Nikita and Vedaldi, Andrea and Rupprecht, Christian and Novotny, David},
    title     = {VGGT: Visual Geometry Grounded Transformer},
    booktitle = CVPR,
    year      = {2025},
    pages     = {5294-5306}
}

@InProceedings{Jiacong2024,
  title={Wild-GS: Real-Time Novel View Synthesis from Unconstrained Photo Collections},
  author={Jiacong Xu and Yiqun Mei and Vishal M. Patel},
  booktitle=NIPS,
  year={2024}
}

@inproceedings{Dahmani2024,
author = {Dahmani, Hiba and Bennehar, Moussab and Piasco, Nathan and Rold\~{a}o, Luis and Tsishkou, Dzmitry},
title = {SWAG: Splatting in the Wild Images with Appearance-Conditioned Gaussians},
year = {2024},
booktitle = ECCV,
pages = {325–340},
numpages = {16},
}

@article{Vandermaaten2008,
  author  = {Laurens van der Maaten and Geoffrey Hinton},
  title   = {Visualizing Data using t-SNE},
  journal = {Journal of Machine Learning Research},
  year    = {2008},
  volume  = {9},
  number  = {86},
  pages   = {2579--2605},
}

@article{Shazeer2020,
title   = {GLU Variants Improve Transformer},
author  = {Shazeer, Noam},
journal = {arXiv preprint arXiv:2002.05202},
year    = {2020}
}

@article{Su2021,
title   = {RoFormer: Enhanced Transformer with Rotary Position Embedding},
author  = {Su, Jianlin and Lu, Yu and Pan, Shengfeng and Wen, Bo and Liu, Yunfeng},
journal = {arXiv preprint arXiv:2104.09864},
year    = {2021}
}

@article{Fujimura2025,
    author    = {Fujimura, Yuki and Kushida, Takahiro and Kitano, Kazuya and Funatomi, Takuya and Mukaigawa, Yasuhiro},
    title     = {UFV-Splatter: Pose-Free Feed-Forward 3D Gaussian Splatting Adapted to Unfavorable Views},
    journal   = {arXiv preprint arXiv:2507.22342},
    year      = {2025}
}

\newpage

\appendix

{\Large {\bf Appendix}}

\section{Training dataset}\label{sec:appendix_training_dataset}

We use the MegaScenes dataset~\cite{Tung2024}, which consists of Internet photos of famous landmarks worldwide.
Although each scene contains multi-view images with camera parameters estimated by COLMAP~\cite{Schonberger2016}, the images are unstructured, requiring the construction of view sets with sufficient overlap for training.

To compute overlap between images, we utilize depth maps.
Specifically, we first estimate monocular depth maps for all images using Depth Anything 3~\cite{depthanything3}.
These depth maps are then aligned with COLMAP sparse depth by estimating scale and shift via RANSAC.

We first select a seed view $\mathbf{I}_1 \in \mathbb{R}^{H \times W \times 3}$.
We then warp the aligned depth map of the seed view $\widetilde{\mathbf{D}}_1$ to another view $\mathbf{I}_i$ and compute the coverage as
\begin{equation}
\mathbf{M}_{1 \to i}(p) = 
\begin{cases}
1 & \text{if } 
| \log \widetilde{\mathbf{D}}_i(\pi_{1 \to i}(p)) - \log \widetilde{\mathbf{D}}_{1 \to i}(p) | < \delta, \\
0 & \text{otherwise},
\end{cases}
\end{equation}
\begin{equation}
\text{CoV}_{1 \to i} = \frac{\sum_{p \in \Omega_1^s} \mathbf{M}_{1 \to i}(p)}{|\Omega_1^s|},
\end{equation}
where $\Omega_1^s$ denotes the sky region computed from the sky probability map of Depth Anything 3, and $|\Omega_1^s|$ is the number of pixels in the sky region.
The warping process is similar to the visibility mask defined in Eq.~(\ref{eq:mask}), and we use the same threshold $\delta = 0.05$.
We also compute the reverse direction $\text{CoV}_{i \to 1}$ and define the coverage score between the two views as
\begin{equation}
\text{CoV}_i = \min(\text{CoV}_{1 \to i}, \text{CoV}_{i \to 1}).
\end{equation}
To ensure sufficient overlap, we select views $\mathbf{I}_i$ with $\text{CoV}_i > 0.5$ and use these two views as context views.

To select a target view, we first choose an interpolated view $\mathbf{I}_j$ between the two context views $\mathbf{I}_1$ and $\mathbf{I}_2$ that satisfies
\begin{equation}
(d_{1j} < d_{12}) \land (d_{j2} < d_{12}) \land (\theta_{1j} < \theta_{12}) \land (\theta_{j2} < \theta_{12}),
\end{equation}
where $d$ and $\theta$ denote the distance and angular difference between two views, respectively.
We then compute the visibility mask at $\mathbf{I}_j$ in the same manner as Eq.~(\ref{eq:mask}).
Finally, we select views with visibility greater than 90\%, i.e., $\sum_{p \in \Omega_j^s} \mathbf{M}_j(p) / |\Omega_j^s| \geq 0.9$, as target views, ensuring sufficient overlap and reducing the influence of transient objects.

Each training sample therefore consists of two context views and one target view ($N_c = 2$, $N_t = 1$), as shown in Fig.~\ref{fig:dataset}.
Although the original MegaScenes dataset contains 458K scenes, our strict filtering results in 14,816 view sets from 3,634 scenes.

\section{Implementation details}

\subsection{Architecture of appearance estimation module}

Our appearance estimation module consists of two ViT blocks, an MLP, and output convolutional layers.
Each ViT block includes a self-attention layer and a feed-forward network with SwiGLU activation~\cite{Shazeer2020}.
We use RoPE as positional encoding~\cite{Su2021} in each block.
The MLP is a standard three-layer network with GeLU activation.

The output convolutional layers are initialized from the DPT head of Depth Anything 3.
In the original DPT head, input images are injected via CNN layers.
In contrast, we omit this module to prevent the model from directly encoding appearance information from the input images.

The additional channels for the appearance embedding are initialized to zero.
We set the dimension of the appearance embedding to $d_g = 32$.

\subsection{Training details}
For the training loss, we set $\lambda = 0.5$.
We use AdamW~\cite{Loshchilov2019} with $\beta_1 = 0.9$, $\beta_2 = 0.95$, and a weight decay of 0.05.
We use a learning rate of $5.0 \times 10^{-5}$, except for the depth offset output layers, for which we use $1.0 \times 10^{-5}$.
The first 2K iterations are warm-up steps, after which we decrease the learning rate using a cosine annealing schedule.
Training is performed for 15K iterations on four NVIDIA A100 GPUs with a per-GPU batch size of 16, and completes in approximately two days.

\subsection{Alignment of 3D Gaussians}

During training, we align the scale of 3D Gaussians from the model to the dataset scale to render the target images, following previous work~\cite{Fujimura2025}.
Assuming the first cameras for both the model output and the dataset are canonical, we estimate a scale factor and translation between point clouds $\{\mathbf{P}_i\}_{i=1}^{2}$ and $\{\widetilde{\mathbf{P}}_i^c\}_{i=1}^{2}$, derived from the output depth $\{\mathbf{D}_i\}_{i=1}^{2}$ and dataset depth $\{\widetilde{\mathbf{D}}_i^c\}_{i=1}^{2}$.
Note that we assume two context views ($N_c = 2$), as described in Sec.~\ref{sec:appendix_training_dataset}.

We estimate a scale factor $a \in \mathbb{R}$ and a translation vector $\mathbf{b} \in \mathbb{R}^3$ by minimizing the weighted least squares error:
\begin{equation}\label{eq:wls}
\min_{a,\mathbf{b}} \sum_{i=1}^{2} \sum_{p \in \Omega} \mathbf{W}_i(p)\left\| a
\mathbf{P}_i(p)
+ \mathbf{b} - \widetilde{\mathbf{P}}^c_i(p)
\right\|^2,\quad \Omega = [1,H] \times [1,W].
\end{equation}

The weight $\mathbf{W}_i(p)$ accounts for errors in monocular depth estimation, and is computed based on geometric consistency between the two context views:
\begin{equation}
\mathbf{W}_1(p) = 
\begin{cases}
\exp \left\{- \gamma| \log \widetilde{\mathbf{D}}_2^c(\pi_{1 \to 2}(p)) - \log \widetilde{\mathbf{D}}_{1 \to 2}^c(p) | \right\} & \text{if } p \in \Omega_1^s, \\
0 & \text{otherwise},
\end{cases}
\end{equation}
where we set $\gamma = 10$.
$\mathbf{W}_2(p)$ is computed in the same manner.

Eq.~\ref{eq:wls} can be solved in closed form as
\begin{equation}
\begin{bmatrix}
a \\
\mathbf{b}
\end{bmatrix}
= \left(\sum_{i=1}^{2} \sum_{p \in \Omega} \mathbf{W}_i(p) \mathbf{X}_i(p)^\top \mathbf{X}_i(p) \right)^{-1} \left( \sum_{i=1}^{2} \sum_{p \in \Omega} \mathbf{W}_i(p) \mathbf{X}_i(p)^\top \widetilde{\mathbf{P}}^c_i(p)\right),
\end{equation}
where $\mathbf{X}_i(p) = [\mathbf{P}_i(p), \mathbf{I}] \in \mathbb{R}^{3 \times 4}$ and $\mathbf{I} \in \mathbb{R}^{3 \times 3}$ is the identity matrix.

Using the estimated $a$ and $\mathbf{b}$, we transform the Gaussian centers and scales as follows:
\begin{align}
\boldsymbol{\mu}_i & \leftarrow a \boldsymbol{\mu}_i + \mathbf{b}, \\
\mathbf{s}_i & \leftarrow a \mathbf{s}_i.
\end{align}

After computing $a$ and $\mathbf{b}$, we estimate the geometric error using Eq.~(\ref{eq:wls}).
During training, we discard samples whose errors exceed 0.5 to avoid adverse effects from unreliable data.

\begin{table}
\caption{Ablation study on the dimension of the appearance embedding.
We report PSNR ($\uparrow$) and LPIPS ($\downarrow$).
The best results are highlighted in red.}
\label{tab:ablation_study_app_embed}
\centering
\begin{adjustbox}{width=\textwidth}
\begin{tabular}{lcc|cc|cc|cc|cc|cc}
\toprule
Scene & \multicolumn{2}{c}{{\it europa}} & \multicolumn{2}{c}{{\it lk2}} & \multicolumn{2}{c}{{\it lwp}} & \multicolumn{2}{c}{{\it schloss}} & \multicolumn{2}{c}{{\it st}} & \multicolumn{2}{c}{{\it stjohann}} \\
\midrule
& PSNR & LPIPS & PSNR & LPIPS & PSNR & LPIPS & PSNR & LPIPS & PSNR & LPIPS & PSNR & LPIPS  \\
\midrule
$d_g=256$ & 15.51 & 0.427 & \cellcolor{bestcolor}16.50 & 0.381 & \cellcolor{bestcolor}13.94 & 0.465 & 16.60 & 0.349 & 14.48 & \cellcolor{bestcolor}0.437 & \cellcolor{bestcolor}13.81 & \cellcolor{bestcolor}0.381 \\
$d_g=32$ & \cellcolor{bestcolor}16.00 & \cellcolor{bestcolor}0.421 & 16.37 & \cellcolor{bestcolor}0.375 & 13.81 & \cellcolor{bestcolor}0.455 & \cellcolor{bestcolor}17.41 & \cellcolor{bestcolor}0.334 & \cellcolor{bestcolor}14.59 & \cellcolor{bestcolor}0.437 & 13.66 & 0.384 \\
\bottomrule
\end{tabular}
\end{adjustbox}
\end{table}

\begin{figure}[t]
\centering
\includegraphics[width=0.8\textwidth]{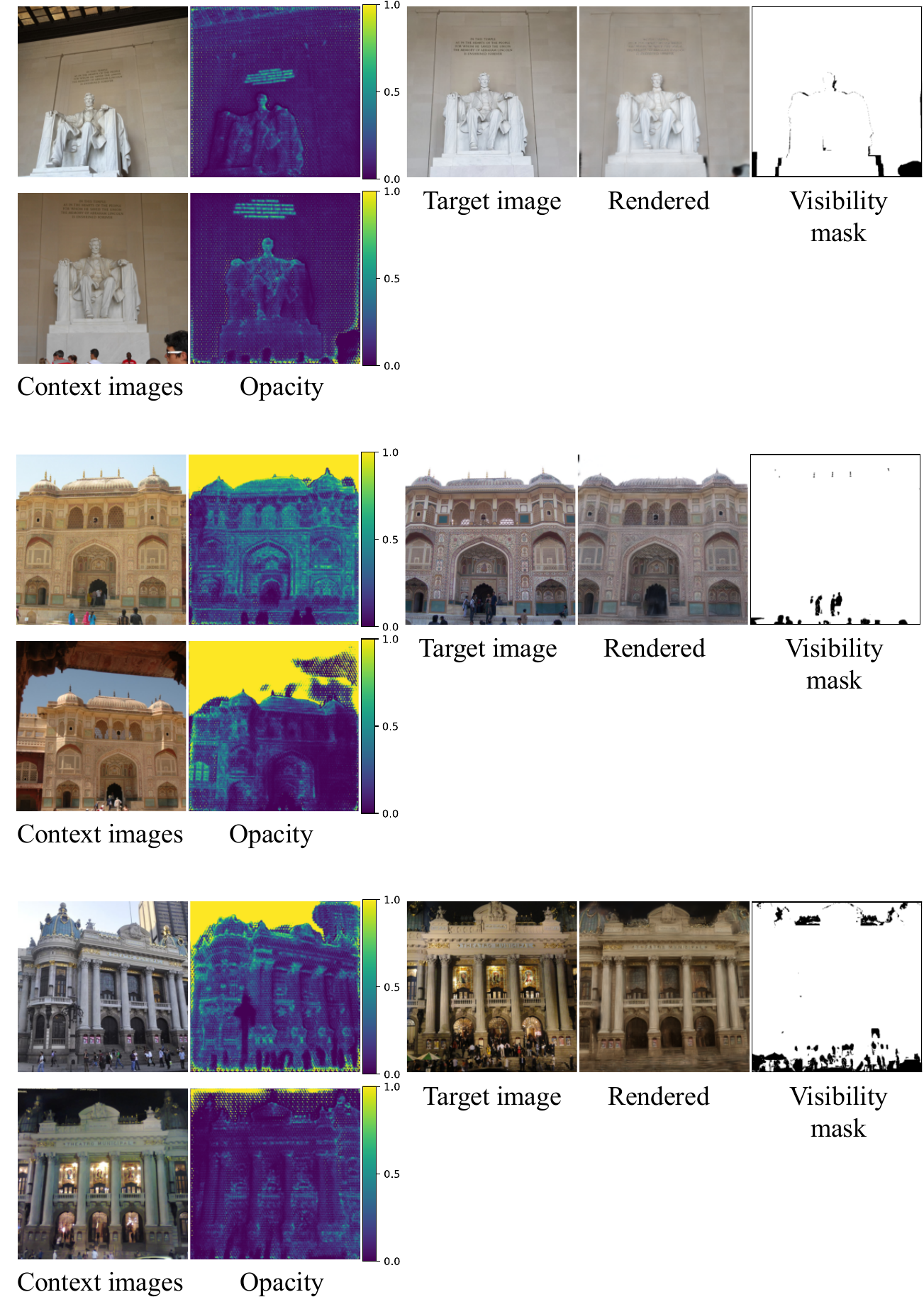}
\caption{
From left to right: context images, estimated opacity maps for the context views, target image, rendered target image, and the visibility mask at the target view. The model suppresses transient objects by assigning low opacity values.
}
\label{fig:opacity_vis}
\end{figure}

\section{Additional results}

\subsection{Ablation study on the dimension of the appearance embedding}

Table~\ref{tab:ablation_study_app_embed} presents an ablation study on the dimension of the appearance embedding.
We hypothesize that appearance variations can be effectively modeled in a low-dimensional latent space.
We compare embedding dimensions of $d_g = 32$ and $d_g = 256$, and observe comparable performance across all scenes.
This result indicates that a low-dimensional latent space is sufficient to capture global appearance variations.

\subsection{Learning to remove transient objects}

Our model is trained on in-the-wild photo collections with varying lighting conditions.
Such images, however, often contain transient objects such as pedestrians.
Although our primary goal is to model appearance variation rather than explicitly handle transient objects, our model implicitly learns to suppress them during training.

As described in Sec.~\ref{sec:training_dataset}, our method uses visibility masks for target views, which model occlusions based on geometric consistency, to compute the training loss.
On the other hand, we do not explicitly model transient objects in the context views.

Figure~\ref{fig:opacity_vis} shows examples of training samples, including two context images, a target image, and the corresponding visibility mask extended with sky regions.
We also visualize the learned opacity maps and the rendered target images.
The rendered image does not contain transient objects, as these regions are masked during training.
Interestingly, transient objects in the context images are also suppressed in the final rendering.
This is because the model automatically assigns low opacity values to such regions in the context views.

In this work, we do not explicitly model transient objects in the context views.
An interesting direction for future work is to further analyze this behavior and leverage it for more robust and stable training.

\end{document}